\def\year{2022}\relax
\documentclass[letterpaper]{article} 
\usepackage{aaai22}  
\usepackage{times}  
\usepackage{helvet}  
\usepackage{courier}  
\usepackage[hyphens]{url}  
\usepackage{graphicx} 
\usepackage{natbib}  
\usepackage{caption} 
\DeclareCaptionStyle{ruled}{labelfont=normalfont,labelsep=colon,strut=off} 
\usepackage{amssymb}
\usepackage{subcaption}
\usepackage[normalem]{ulem}
\usepackage{multirow}
\usepackage{algorithm}
\usepackage{algorithmic}

\usepackage{newfloat}
\usepackage{listings}
\floatstyle{ruled}
\newfloat{listing}{tb}{lst}{}
\floatname{listing}{Listing}
\title{Isomer: Transfer enhanced Dual-Channel Heterogeneous Dependency Attention Network for Aspect-based Sentiment Classification}
\author{
    Yukun Cao\textsuperscript{\rm 1},
    Yijia Tang\textsuperscript{\rm 1},
    Ziyue Wei\textsuperscript{\rm 1},
    ChengKun Jin\textsuperscript{\rm 1},
    Zeyu Miao\textsuperscript{\rm 1},
    Yixin Fang\textsuperscript{\rm 1},
    Haizhou Du\textsuperscript{\rm 1},
    Feifei Xu\textsuperscript{\rm 1}
}
\affiliations{
    \textsuperscript{\rm 1}ShangHai University of Electric Power \\
    marilyn\_cao@163.com

}

\usepackage{bibentry}

\begin{document}

\maketitle

\begin{abstract}
Aspect-based sentiment classification aims to predict the sentiment polarity of a specific aspect in a sentence. However, most existing methods attempt to construct dependency relations into a homogeneous dependency graph with the sparsity and ambiguity, which cannot cover the comprehensive contextualized features of short texts or consider any additional node types or semantic relation information. To solve those issues, we present a sentiment analysis model named Isomer, which performs a dual-channel attention on heterogeneous dependency graphs incorporating external knowledge, to effectively integrate other additional information. Specifically, a transfer-enhanced dual-channel heterogeneous dependency attention network is devised in Isomer to model short texts using heterogeneous dependency graphs. These heterogeneous dependency graphs not only consider different types of information but also incorporate external knowledge. Experiments studies show that our model outperforms recent models on benchmark datasets. Furthermore, the results suggest that our method captures the importance of various information features to focus on informative contextual words.
\end{abstract}

\section{Introduction}
\begin{figure}[t]
\centering
\includegraphics[width=0.9\columnwidth]{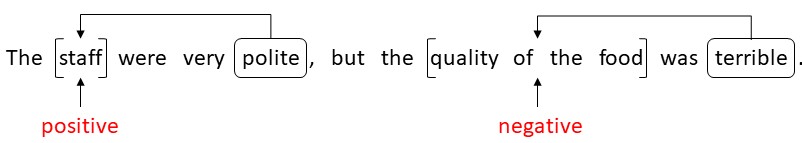}
\caption{An example of a restaurant review with two aspects having different sentiments.}
\label{fig1}
\end{figure}

Sentiment analysis is an important research branch in the field of Natural language processing (NLP), and the purpose of sentiment analysis is to identity the potential sentiment of a text. Aspect-based sentiment analysis (ABSA) is a fine-grained sentiment classification task in the field of sentiment analysis that aims to predict the sentiment polarity (e.g. positive, negative or neutral) of specific aspects in a sentence. As shown in Figure \ref{fig1}, given a restaurant review “The staff were very polite, but the quality of the food was terrible”, ABSA judges the sentiment polarities of the two aspects “staff” and “quality of food” as positive and negative, respectively. Mining aspect-related semantic parsing information and precisely localizing respective opinion words lies at the heart of this task.

Current neural techniques such as RNNs \cite{kiritchenko2014nrc}, CNNs \cite{dong2014adaptive} have already been commonly used in ABSA. To capture the importance between contexts and the given aspects, \cite{xu2015show} and \cite{li2018transformation} respectively proposed RNN-augmented and CNN-augmented methods using an attention mechanism, and both achieved encouraging results in enhancing representations.

Recently, graph convolutional networks (GCNs) \cite{zhang2019aspect} have been used to explore the dependency relations between contexts incorporating the dependency tree structure of a sentence. Most studies mentioned above apply homogeneous dependency graphs and dependency trees to represent contextual syntactic and dependencies. However, only one type of node and a single relation are included in a homogeneous dependency graph, which indicates that those studies merely consider syntactic rules among contextual words.

However, the shortcomings of the above approaches in the ABSA task should not be overlooked. First, using a homogeneous dependency graph to represent text cannot fully cover comprehensive dependencies due to the semantic sparsity and the absent grammer. Second, noisy information is inevitably introduced through the useless information contained in a dependency tree, causing the model to fail to capture the dependencies between aspects and contexts; Third, the contextual features in the model may lead to inaccurate recognition of the relevant information between aspects and contexts due to ambiguity and cannot effectively reflect the differences between contexts, thereby affecting the performance of the model.

To address these three challenges, we propose a novel model Isomer. Isomer first uses external KGs to supplement the background knowledge and emotional knowledge of a text, eliminating the ambiguity of the text and enhancing the representations. Then, we build a context-sentiment oriented heterogeneous dependency graph and a entity-text oriented heterogeneous dependency graph that integrate other types of information. We use these two heterogeneous dependency graphs covering four types of nodes and four types of text relationships to compensate for the defect of single information in homogeneous dependency graphs. Then, the model feds two heterogeneous dependency graphs into the respective transfer-enhanced dual-channel heterogeneous dependency attention network (named Trans\_DHA), captures the importance between node channel features and type channel features and reduces the influence of noise information. Transformer is introduced for iterative interactive learning with the dual-channel heterogeneous dependency attention. Finally, the representations learned from the two heterogeneous dependency graphs are connected so that both the rich relationship between the text and additional information can be mined, and the flat representations and graph-based representations can be jointly considered. We conduct extensive evaluations on 
five benchmark datasets. Experimental results demonstrate that our approach achieves better results compared to other strong competitors.
\section{Related Work}
Most recent research works on ABSA extensively utilize deep learning algorithms, such as CNNs \cite{dong2014adaptive}, RNNs \cite{kiritchenko2014nrc}, and have achieved promising progress. Furthermore, attention-based algorithms have been applied to capture contextual features. Among them, \cite{xu2015show} and \cite{li2018transformation} use an attention mechanism to augmente an RNN and a CNN respectively. Such approaches have achieved promising performance in enhancing various representations. An attention-based neural network \cite{wang2016attention} was proposed to identify important sentiment information related to aspect words. 

In addition, \cite{dong2014adaptive} seek to encode a parse tree (ie. dependency tree) based on dependency relations using a RNN and compute the node distances as attention weight. Recently, graph convolutional networks (GCNs) \cite{zhang2019aspect} have been used to explore the dependencies between contexts by incorporating dependency trees into attention models to improve performance. In another work, \cite{tang2020dependency} propose DGEDT based on aspect and dependency graphs and improve the shortcomings of the instability and noisy information of the dependency tree.

This paper first proposes an isomer framework for processing the heterogeneous dependency graph representation of text. In the framework, we incorporate additional information such as contextual dependencies, emotional words, entities, and sentences in heterogeneous dependency graphs; and the rich relationship between text and additional information can be captured by a transfer-enhanced dual-channel heterogeneous dependency attention network. Moreover, to the best of our knowledge, none of the previous studies are similar to this work, which enhances contextual representations by searching the various relationships between text and foreign knowledge graphs using ConceptNet and SenticNet. Furthermore, we also combine the heterogeneity of sentences to generate heterogeneous dependency graphs integrating multiple additional information for aspect-level sentiment analysis.

\begin{figure*}[t]
\centering
\includegraphics[width=0.9 \textwidth]{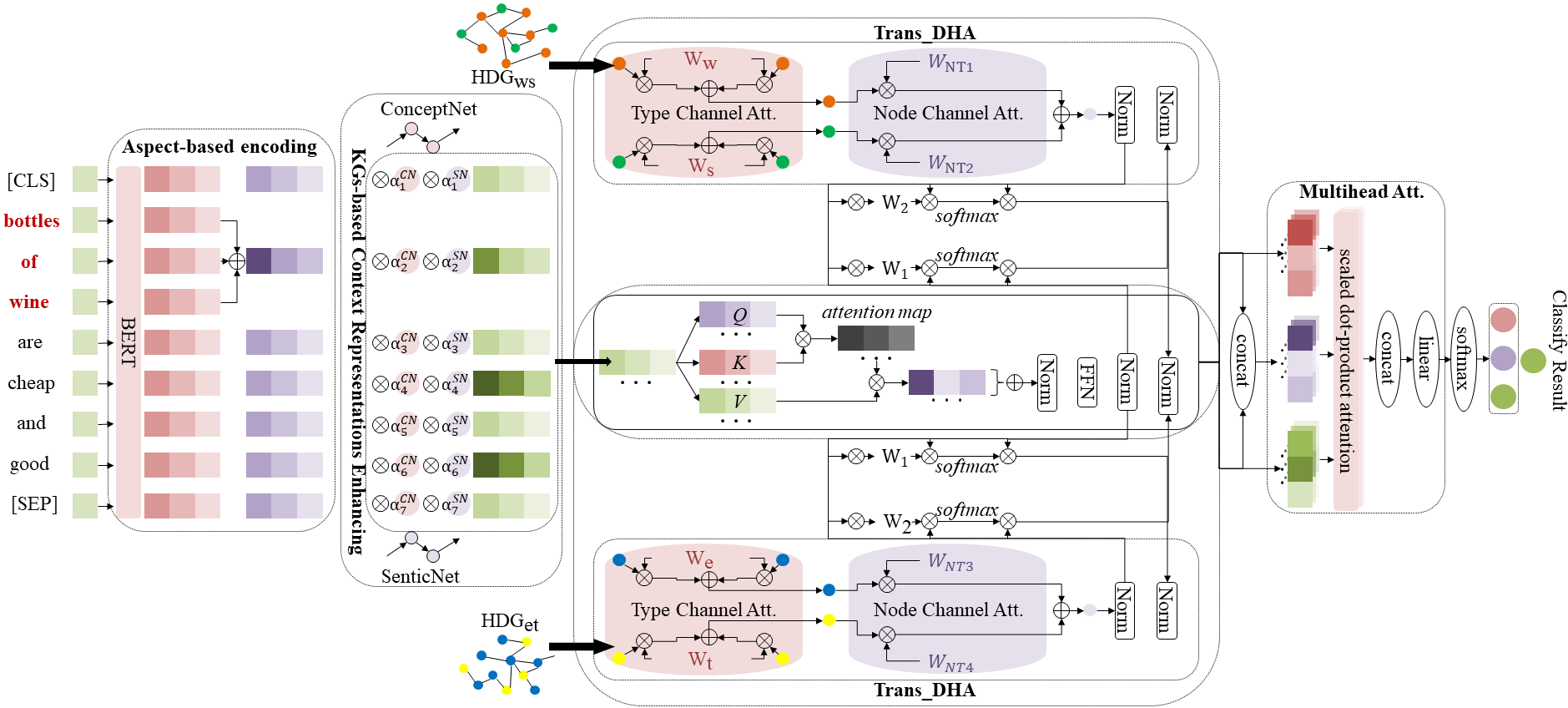}
\caption{An overall demonstration of our proposed Isomer.}
\label{fig2}
\end{figure*}

\section{Isomer}
We first present the ABSA task definitions. Given an n-word sentence ${\rm{S}}=\left [ w_{1},\cdots ,w_{i}^{a},\cdots w_{i+\left ( m-1 \right )}^{a},\cdots ,w_{n} \right ],i\in \left [ 1,n \right ]$ and its corresponding m-word aspects $\left [ w_{i}^{a},\cdots w_{i+\left ( m-1 \right )}^{a} \right ]$. ABSA aims to automatically learn the contextual representations between sentence S and aspects and further judge emotional polarity (e.g., positive, negative and neural).

In this paper, we propose a novel ABSA model called Isomer including two Trans\_DHAs. The entire architecture of our proposed Isomer is shown in Figure \ref{fig2}. The model takes word sequences as input and uses BERT \cite{devlin2018bert} as the encoder to acquire initial sentence representations. Then, we introduce a KG-based context representation-enhancing module, in which the context representations are enhanced under the interaction of the common-sense knowledge graph ConceptNet and the emotional knowledge graph SenticNet. With the cooperation of two heterogeneous dependency graphs considering different types of information, we fed contextual representations into the Trans\_DHA structure. We sent the final representations to the classifier for emotion prediction after further feature fusion.

\subsection{Aspect-Based Encoding}
The model first utilizes BERT as the encoder to obtain the contextual embedding. Specifically, we process the sentence into the token sequence format of BERT as: $\rm \hat{S}=\left [ CLS \right ]+{S}+\left [ SEP \right ]$, where [CLS] and [SEP] are tokens specifically introduced in BERT. The pretrained BERT is applied for the encoder to extract hidden contextual representations ${\rm E}=e_{1},\cdots ,e_{i}^{a},\cdots ,e_{i+\left ( m-1 \right )}^{a},\cdots ,e_{n+2}\left ( i \in \left [ 2,n+1 \right ] \right )$. Here, $e_{i}$ is the embedding of the $i$-th word in token sequence $\widehat{\mathrm{S}}$, and $e_{i}^{a},\cdots ,e_{i+\left ( m-1 \right )}^{a}$ represents the m-word aspects. Considering that the aspects can be one word or a combination of multiple words, we combine each aspect as a single token representation $e_{i}^{a}=e_{i}^{a}+\cdots +e_{i+\left ( m-1 \right )}^{a}$, in order to better judge the relationship between aspects and other tokens in the downstream tasks. Then, we obtain the final contextual representations ${\rm E}=e_{1},\cdots ,e_{i}^{a},\cdots ,e_{n+3-m}$.

\subsection{KG-based Context Representations Enhancement}
KGs usually contain many connections between facts and entities, and these complex relationships in KGs provide a special perspective to explore the potential related information of aspects.

Short texts usually omit much background knowledge, and it is often difficult for us to infer the true sentiment polarity. Therefore, we need to understand the background knowledge of the aspect when judging the emotional orientation of the aspect. We utilize the common-sense knowledge graph ConceptNet \cite{speer2017conceptnet} to supplement its background knowledge. Furthermore, the emotional knowledge graph SenticNet \cite{10.1145/3340531.3412003} is used to promote the tagging of emotional words in the word embedding for the little sentiment information contained in the general word embedding and the requirements for incorporating supernumerary emotional knowledge.

Specifically, we choose aspect $w_{i}^{a}$ as the starting concept node and find the entity set $c_{j}^{{\rm CN}^{o}}$,$j \in \mathbb{R}$ that is related to the aspect in ConceptNet.

\begin{equation}
    c_{w_{i}^{a}}^{\rm CN}\stackrel{\rm retrieve}{\longrightarrow}
    \left \langle c_{w_{i}^{a}}^{\rm CN},{c_{w_{i}^{a},j}^{\rm CN}}^{r},c_{j}^{{\rm CN}^{o}},w_{w_{i}^{a},j} \right \rangle
\end{equation}

\noindent where $c_{w_{i}^{a}}^{\rm CN}$ represents the aspect used to find the entity set, and ${c_{w_{i}^{a},j}^{\rm CN}}^{r}$, $c_{j}^{{\rm CN}^{o}}$ and $w_{w_{i}^{a},j}$  represent all relationships between aspects and entities, target entities and weights, respectively.

We then search for the context $w_{i}$ in sentence S that is related to those entities $c_{j}^{{\rm CN}^{o}}$. Once it exists, the contextual representation is assigned a weight $\alpha _{w_i}^{\rm CN}$, which indicates that this context is closely related to the aspect.
\begin{equation}
    \left \langle c_{j}^{{\rm CN}^{o}},c_{w_{i}}^{\rm CN} \right \rangle \stackrel{\rm retrieve}{\longrightarrow}\left \langle c_{w_{i}}^{\rm CN},{c_{{w_i},j}^{\rm CN}}^r,c_{j}^{{\rm CN}^{o}},w_{w_i,j} \right \rangle
\end{equation}
Note that for general embedding, it is very weak to tag sentiment words such as “good” and “bad” and is not conducive to downstream sentiment analysis tasks. That is, there is very little emotional information contained in the word vector. Based on the above, we first use GI\footnote{http://wjh.harvard.edu/~inquirer/} (General Inquirer), the evaluator dictionary, to determine the sentiment sets $\left\{w_{i}^{s}\right\}$ in sentence S and then extract the entities $\left\{s_{j}^{\rm S N^{o}}\right\}$ in sentence that are related to the sentiment word using the emotional knowledge graph SenticNet.

\begin{equation}
    s_{w_{i}^{s}}^{\rm SN}\stackrel{\rm retrieve}{\longrightarrow}
    \left \langle s_{w_{i}^{s}}^{\rm SN},{s_{w_{i}^{s},j}^{\rm SN}}^{r},s_{j}^{{\rm SN}^{o}},w_{w_{i}^{s},j} \right \rangle
\end{equation}
\noindent where $s_{w_{i}^{s}}^{\rm SN}$ represents the sentiment word; and ${s_{w_{i}^{s},j}^{\rm SN}}^{r}$, $s_{j}^{{\rm SN}^{o}}$ and $w_{w_{i}^{s},j}$ represent all relationships between the context and sentiments, target entities and weights, respectively. We search for context $w_{i}$ in sentence S, which is related to entity $s_{j}^{{\rm SN}^{o}}$; If it exists, the context is assigned an emotional contribution $\alpha_{w_{i}}^{\rm SN}$, which indicates that this context is closely related to the sentiments, and 
the sentiment knowledge is incorporated into embedding.
\begin{equation}
    \left \langle s_{j}^{{\rm SN}^{o}},s_{w_{i}}^{\rm SN} \right \rangle \stackrel{\rm retrieve}{\longrightarrow}\left \langle s_{w_{i}}^{\rm SN},{s_{w_i,j}^{\rm SN}}^r,s_{j}^{{\rm SN}^{o}},w_{w_i,j} \right \rangle
\end{equation}

\subsection{Construction of Heterogeneous Dependency Graph}
Due to the semantic sparsity and the absent grammer of short text, using a homogeneous dependency graph cannot fully cover comprehensive dependency relations. The isomorphic nature of the homogeneous dependency graph makes it impossible to consider other node types and semantic relations in the dependency graph. However, previous studies used external knowledge entities and potential topics in the knowledge base to enrich the semantics of short texts but did not consider sentence dependency. Therefore, we model dependency, entities, sentences and sentiment words using two heterogeneous dependency graphs: a context-sentiment oriented heterogeneous dependency graph and an entity-text oriented heterogeneous dependency graph.

We construct a context-sentiment oriented heterogeneous dependency graph ${\rm HDG}_{ws}$ containing contextual words and sentiments. The set of edges E consists of the connections between contextual words and sentiments. Similar to the entity-text oriented heterogeneous dependency graph ${\rm HDG}_{et}$, we build it with entities and texts. The set of edges E represents two relations: relationships among entities and relations between entities and sentence. 

\subsection{Trans\_DHA: Transfer-enhanced Dual-Channel Heter-ogeneous Dependency Attention Network}
A homogeneous dependency graph is adopted by most modern studies to represent a dependency tree. The nodes of the graph consist of the words in sentences, and the edges consist of the dependency information. In other words, there is only one node type and one edge type in such a graph. Besides, most research will use GCNs to address dependency graph structures. However, the traditional GCN is a multilayer neural network that operates in a straightforward way on a homogeneous graph. If a GCN for homogeneous graphs is simply used on heterogeneous graphs, considerable information in the graphs, including some structural information and meta-paths that can reflect the hidden semantic information in heterogeneous graphs, will be lost. Therefore, to use a GCN to process heterogeneous graphs, it is necessary to solve the problem that GCNs cannot handle the feature space of different types of nodes.

To this end, we propose a transfer-enhanced dual-channel heterogeneous dependency attention network called Trans\_DHA, and its main idea is to consider the difference of the information about two different types of nodes in heterogeneous dependency graph and use two projection matrices for different nodes to project them into a vector space, which solves the problem that a traditional GCN cannot effectively handle heterogeneous graphs. Meanwhile Transformer is introduced for iterative interactive learning with dual-channel heterogeneous dependency attention. Finally, the representations learned from the two heterogeneous dependency graphs are connected so that both the rich relationships between the text and additional information can be mined.
\begin{equation}
    {h^{l+1}}^{k}= {\rm Relu} \left( \sum_{\tau }\tilde {\tilde {A_{\tau}^{k}}} h_{\tau }^{l}  W_{\tau }^l \right)
\end{equation}
\noindent where $\tilde {\tilde {A_{\tau}^{k}}}$ represents the symmetric normalized adjacency matrix of heterogeneous dependency graph $k$, and the node type is $\tau$. The representation of the nodes $h^{l+1}$ is obtained by aggregating all their neighboring nodes representations $h^{l}$ using different projection matrices $W_{\tau}^{l} \in \mathbb{R}^{d^{l} \times d^{l+1}}$ for different types $\tau$ of nodes. The projection matrix $W_{\tau}^{l}$ considers the difference of different types of node representations and projects them into a public feature space $\mathbb{R}^{d^{l+1}}$ to perform feature fusion.

\subsubsection{Type Channel Attention}
We utilize the type channel attention to learn the scores of different types of neighboring nodes of a specific node $i$. The node feature is defined as the sum of the neighboring node features. Then, we calculate the type channel attention scores and the type channel attention weight will be obtaioned with the softmax function to normalize the attention scores.

\subsubsection{Node Channel Attention}
We design node channel attention to capture the importance of different neighboring nodes. Given a node $i$ whose type is $\tau$ and its neighboring node $j \in N_{i}$ with type $\tau'$, the node channel attention scores and the node channel attention weight can be computed based on the node features and the type channel attention weight of node $j$.

\subsubsection{Iterative Interactive Learning}
After introducing type channel attention and node channel attention into the dual-channel heterogeneous dependency attention structure, we can thus obtain two heterogeneous dependency graph based representations of input text. In addition, we also use the introduced Transformer and dual-channel heterogeneous dependency attention to perform iterative interactive learning through a mutual transformation process to further enhance the text representations $h$.

\subsection{Loss function}
The final representation h mapped to probabilities over the sentiment polarities is optimized by the standard gradient descent algorithm with the cross-entropy loss and L2-regularization:

\begin{equation}
   L=-\sum_{\left ( d,y \right )\in {\rm D}} \log\left ( {\rm softmax} \left(h\right) \right )+\lambda \left \| \vartheta  \right \|
\end{equation}

\noindent where D denotes the training dataset, and $y$ is the ground-truth label. $\vartheta$, represents all trainable parameters, and $\lambda$ is the coefficient of the regularization term.

\section{Experiments}

\subsection{Datasets}

\begin{table}[]
\centering
\begin{tabular}{c|c|c|c|c}
\hline
\multicolumn{2}{c|}{Dataset}     & \#Pos. & \#Neu. & \#Neg. \\ \hline
\multirow{2}{*}{Twitter} & Train & 1561   & 3127   & 1560   \\
                         & Test  & 173    & 346    & 173    \\ \hline
\multirow{2}{*}{Lap14}   & Train & 994    & 464    & 870    \\
                         & Test  & 341    & 169    & 128    \\ \hline
\multirow{2}{*}{Rest14}  & Train & 2164   & 637    & 807    \\
                         & Test  & 728    & 196    & 196    \\ \hline
\multirow{2}{*}{Rest15}  & Train & 912    & 36     & 256    \\
                         & Test  & 326    & 34     & 182    \\ \hline
\multirow{2}{*}{Rest16}  & Train & 1240   & 69     & 436    \\
                         & Test  & 469    & 30     & 117    \\ \hline
\end{tabular}
\caption{Datasets statistics.}
\label{table1}
\end{table}

\begin{table*}[]
\centering
\begin{tabular}{c|c|c|c|c|c|c|c|c|c|c}
\hline
\multirow{2}{*}{Model} &
  \multicolumn{2}{c|}{Twitter} &
  \multicolumn{2}{c|}{Lap14} &
  \multicolumn{2}{c|}{Rest14} &
  \multicolumn{2}{c|}{Rest15} &
  \multicolumn{2}{c}{Rest16} \\ \cline{2-11} 
                 & Acc         & F1          & Acc   & F1     & Acc   & F1     & Acc  & F1     & Acc  & F1     \\ \hline
AOA\cite{huang2018aspect}        & 70.2        & 67.8        & 71.9  & 66.0   & 77.7  & 63.2   & 77.9 & 52.3   & 84.9 & 52.8   \\ \hline
TNETLF\cite{li2018transformation}   & 70.2        & 68.7        & 72.4  & 66.3   & 80.1  & 70.9   & 78.0 & 52.2   & 87.7 & 57.9   \\ \hline
ASGCN\cite{zhang2019aspect}      & 72.5        & 70.8        & 75.7  & 71.6   & 81.3  & 72.7   & 79.1 & 61.3   & 88.2 & 69.8   \\ \hline
ASCNN\cite{zhang2019aspect}      & 72.5        & 70.6        & 75.0  & 71.2   & 81.2  & 72.8   & 80.4 & 60.8   & 88.0 & 67.9   \\ \hline
BIGCN\cite{zhang2020convolution}      & 72.7        & 71.3        & 74.9  & 70.6   & 80.8  & 71.8   & 80.4 & 64.8   & 89.0 & 70.7   \\ \hline
DGEDT\cite{tang2020dependency} &
  76.5 &
  75.5 &
  \underline{78.8} &
  \underline{75.3} &
  \underline{86.1} &
  \underline{80.0} &
  \underline{82.9} &
  \underline{68.7} &
  \underline{91.7} &
  \underline{77.2} \\ \hline
Dual\_MRC\cite{mao2021joint}  & -           & -           & -     & 64.6 & -     & 76.6 & -    & 65.1 & -    & 70.8 \\ \hline
DualGCN\cite{li-etal-2021-dual-graph}    & 75.9       & 74.3       & 78.5 & 74.7  & 84.3 & 78.1  & -    & -      & -    & -      \\ \hline
\textbf{Isomer} &
  \textbf{78.3} &
  \textbf{77.6} &
  \textbf{79.9} &
  \textbf{76.5} &
  \textbf{87.2} &
  \textbf{80.9} &
  \textbf{85.5} &
  \textbf{72.8} &
  \textbf{92.5} &
  \textbf{80.0} \\ \hline
\end{tabular}
\caption{Comparison results for all methods in terms of accuracy and F1 score.“-” means not reported. We show the results of our model (Isomer) in the last row. The best results on each dataset are in bold. The second best results are underlined.}
\label{table2}
\end{table*}

Our experiments are conducted on five datasets, as shown in Table \ref{table1}. The first dataset (Twitter) was built by \cite{dong2014adaptive}. The other four datasets (Lap14, Rest14, Rest15 and Rest16) are from SemEval 2014 task 4 \cite{pontiki-etal-2014-semeval}, SemEval 2015 task 12 \cite{pontiki2015semeval} and SemEval 2016 task 5 \cite{hercig2016uwb}, which contain reviews on laptops and restaurants.

\subsection{Implementation Details}
We initialize word embeddings using the 300-dimensional GloVe vectors provided by \cite{pennington2014glove} and report the average maximum value for all metrics on the testing set. We use the BERT-base English version with Adam \cite{kingma2014adam} as the optimizer, and the learning rate is set as 0.001. We set the hidden size of Isomer to 512. As for regularization, a dropout function is applied to word embeddings, and the dropout rate is set as 0.3. Besides, the coefficient $\lambda$ for the L2-regularization is set as 0.0001, and the batch size is 32. Accuracy and Macro-Averaged F1 are used as the evaluation metrics.

\subsection{Baseline Methods}
To comprehensively evaluate Isomer for sentiment classification, we compare it with the following state-of-the-art methods:

(1) DualGCN: The DualGCN considers both syntactic structure and semantic relevance, and the two graph convolutional networks play a complementary role in capturing features.

(2) DualMRC: The DualMRC proposes a joint training dual-MRC framework to handle all subtasks of aspect-based sentiment analysis in one shot.

(3) DGEDT: The DGEDT proposes a dual-transformer structure that considers the connections in the dependency tree as a supplementary GCN module to diminish the error induced by incorrect dependency trees.

(4) BiGCN: The BiGCN is a graph-based method for aspect-level sentiment classification tasks. It employs both the ordinary syntactic graph and a lexical graph to capture the global word co-occurrence information in a cooperative way.

(5) ASGCN: The ASGCN constructs a dependency tree using syntactical information and word dependencies.

(6) ASCNN: The ASCNN replaces the 2-layer GCN in the ASGCN with a 2-layer CNN.

(7) TNETLF: TNetLF uses a CNN layer instead of an attention layer to extract the relevant features from the word representations generated by the bidirectional RNN.

(8) AOA: The AOA uses multiple attention layers to simulate the interaction between aspects and sentences.

(9) Isomer: Our approach.

\subsection{Experimental Results}
Table \ref{table2} shows the classification accuracies and F1 scores of different methods on 5 benchmark datasets. We see that the performance of our model (Isomer) outperforms recent models, including the dependency tree-based models (ASGCN, ASCNN, DGEDT), the syntactic and semantic-based models (BiGCN, DualGCN), the attention-based models (AOA, DualMRC, TNETLF), in both accuracy and F1 score. The main reason is Isomer considers multiple types of information, including dependencies, entities, sentences and sentiment words. In addition, a newly proposed dual-channel attention embeds different types of features into the semantics of text. Our Isomer model also superimposes external knowledge to enhance contextual semantics, and its results have also achieved significant improvements. The above results prove the effectiveness of Isomer at performing aspect-based sentiment analysis and capturing important syntax, grammatical structure and emotional knowledge. 

\subsection{Ablation Study}
In order to verify the validity of Isomer, we investigate and report the results of six typical ablation conditions. The results are shown in Table \ref{table3}.

\begin{table}[H]
\begin{tabular}{c|c|c|c|c|c}
\hline
\multirow{2}{*}{Ablation} & Twitter       & Lap14         & Rest14        & Rest15        & Rest16        \\ \cline{2-6} 
                          & Acc           & Acc           & Acc           & Acc           & Acc           \\ \hline
\textbf{Isomer}                    & \textbf{78.3} & \textbf{79.9} & \textbf{87.2} & \textbf{85.5} & \textbf{92.5} \\ \hline
w/o ws          & 76.7          & 78.7          & 85.2          & 83.1          & 92.0          \\ \hline
w/o et          & 76.9          & 78.0          & 86.5          & 83.3          & 92.0          \\ \hline
w/o Hete                  & 76.6          & 77.3          & 85.4          & 83.0          & 91.2          \\ \hline
w/o CN            & 77.9          & 78.5          & 85.7          & 85.3          & 92.4          \\ \hline
w/o SN             & 77.3          & 79.6          & 85.6          & 84.3          & 92.0          \\ \hline
w/o KGs                   & 77.1          & 78.0          & 85.2          & 83.7          & 91.9          \\ \hline
\end{tabular}
\caption{Overall ablation results of the accuracy on five datasets. ‘CN’ and ‘SN’ respectively represent ConceptNet and SenticNet, ‘w/o’ denotes without.}
\label{table3}
\end{table}

We first studied the influence of the proposed Trans\_DHA. Compared with the complete Isomer, the performance of one Trans\_DHA (w/o et and w/o ws) is reduced, indicating that using two heterogeneous dependency graphs at the same time has better performance because it adds more types of information. Then, we further remove the two Trans\_DHAs simultaneously (w/o Hete) and observe that do not utilizing this structure is worse than Isomer. This clearly reveals the positive impact of our proposed Trans\_DHA that consider the heterogeneity of multiple information, and  aggregates different types of features to the embedding vectors.

Finally, we also studied the importance of the external KGs. Here, w/o CN, w/o SN and w/o KGs respectively indicate the removal of ConceptNet, the removal of the SenticNet and the removal of the two knowledge graphs at the same time. The results show that the use of a common-sense knowledge graph and emotional knowledge graph can enhance the contextual representation. And the result on Rest14 indicates that using two knowledge graphs at the same time allows the two types of knowledge to complement to each other.

\section{Conclusion}
In this paper, we emphasize the importance of different semantic rules in the ABSA task. In order to use these rules, we propose a heterogeneous framework Isomer that uses heterogeneous dependent attention. First, Isomer eliminates the ambiguity of the short text by introducing external KGs. Different from most current studies using homogeneous dependency graphs merely containing a single semantic relationship, we use two heterogeneous dependency graphs that integrate other types of information to capture the global interactive information of the text. Besides, we have also added a transfer-enhanced dual-channel heterogeneous dependency attention network, which can capture different types of node characteristics and node relationships in heterogeneous dependence graphs. The results on five datasets demonstrate that our model indeed promotes the final performance and achieves state-of-the-art performance.

\bibliography{Isomer}
\end{document}